\documentclass[11pt]{article}

\usepackage{emnlp2021}
\usepackage{graphicx}

\usepackage{times}
\usepackage{latexsym}

\usepackage{xcolor, soul}
\usepackage{graphicx}
\usepackage{booktabs}

\definecolor{lightgreen}{rgb}{0.56, 0.93, 0.56}
\definecolor{moonstoneblue}{rgb}{0.45, 0.66, 0.76}

\definecolor{teal}{HTML}{0B8494}
\definecolor{orange}{HTML}{FFBE98}
\definecolor{purple}{HTML}{F05A7E}
\definecolor{blue}{HTML}{0096FF}

\usepackage[most]{tcolorbox}
\tcbset{on line, 
        boxsep=4pt, left=0pt,right=0pt,top=0pt,bottom=0pt,
        colframe=white,colback=teal,  
        highlight math style={enhanced}
        }

\usepackage{amssymb}
\usepackage{pifont}

\newcommand{\middleschool}{\tcbox{\color{white}M}}%
\newcommand{\highschool}{{\tcbox[colback=orange]{\color{white}{H}}}}%
\newcommand{\college}{{\tcbox[colback=purple]{\color{white}{C}}}}%
\newcommand{\elementary}{{\tcbox[colback=blue]{\color{white}{E}}}}%

\usepackage[T1]{fontenc}

\usepackage[utf8]{inputenc}

\usepackage{microtype}
\usepackage{svg}
\usepackage{epigraph}
\usepackage{multirow}
\usepackage{etoolbox}
\usepackage{scalerel,xparse}

\newcommand{\emojiro}{\scalerel*{\includegraphics{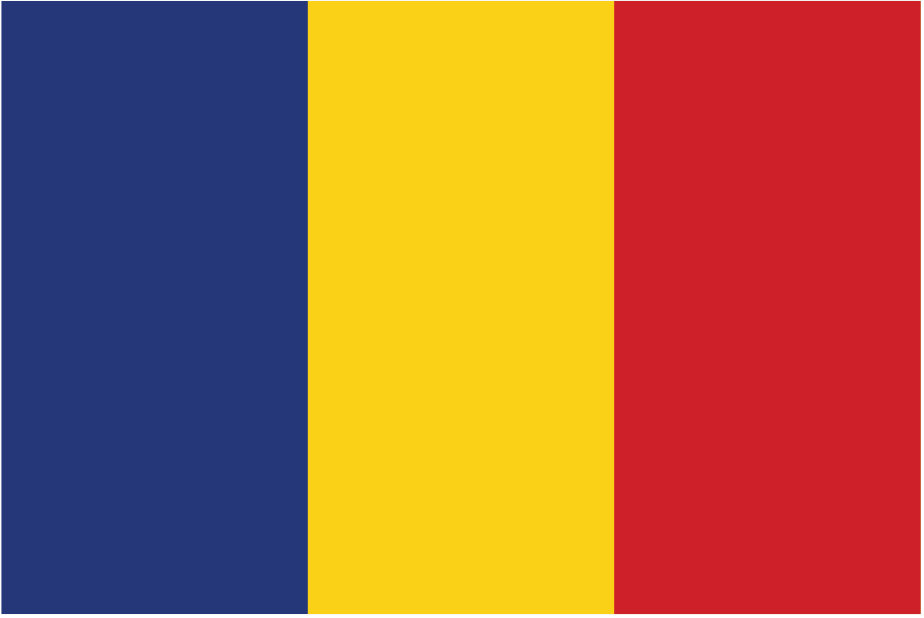}}{X}}

\title{RoMath: A Mathematical Reasoning Benchmark in Romanian}

\author{Adrian Cosma$^1$, Ana-Maria Bucur$^2$, Emilian Radoi$^1$ \\
    {$^{1}$POLITEHNICA Bucharest National University of Science and Technology} \\
    {$^2$Interdisciplinary School of Doctoral Studies, University of Bucharest} \\
    {București, Romania} \\
    {\small \texttt{\{ioan\_adrian.cosma, emilian.radoi\}@upb.ro}}\\
    \small \texttt{ana-maria.bucur@drd.unibuc.ro}\\}

\begin{document}
\maketitle


\begin{abstract}
Mathematics has long been conveyed through natural language, primarily for human understanding. With the rise of mechanized mathematics and proof assistants, there is a growing need to understand informal mathematical text, yet most existing benchmarks focus solely on English, overlooking other languages. This paper introduces RoMath, a Romanian mathematical reasoning benchmark suite comprising three subsets: \textit{Baccalaureate}, \textit{Competitions} and \textit{Synthetic}, which cover a range of mathematical domains and difficulty levels, aiming to improve non-English language models and promote multilingual AI development. By focusing on Romanian, a low-resource language with unique linguistic features, RoMath addresses the limitations of Anglo-centric models and emphasizes the need for dedicated resources beyond simple automatic translation. We benchmark several open-weight language models, highlighting the importance of creating resources for underrepresented languages. Code and datasets will be made available.
\end{abstract}

\epigraph{"Matematica s-o fi scriind cu cifre dar poezia nu se scrie cu cuvinte."\footnotemark}{\textit{Nichita Stanescu, "Matematica poetica"}, \\Poem dedicated to mathematician Solomon Marcus.}
\footnotetext{English translation: \textit{"Mathematics may be written with numbers, but poetry is not written with words."}}

\section{Introduction}
Mathematics has been a central intellectual preoccupation to humans since the beginning of civilization, the first mathematical writings dating back approximately 4000 years \cite{FRIBERG1981277}. Historically and in the present, mathematics has been mostly written, spoken and taught in natural language, albeit with its own specialized vocabulary, having strict formalism only sparsely introduced between free-text explanations and reasoning. The primary audience of mathematical reasoning is other humans, not computers. The natural language of mathematics contains a mix of formulas, symbols, neologisms, jargon and words with different meanings than their common meaning (e.g., "real" / "imaginary" numbers). Mathematics implies rigor and precise reasoning, qualitatively different from general NLP. There is a pressing need to automatically process and understand the existing large amount of mathematical text written in natural language to enable efficient knowledge extraction, facilitate automated theorem proving, and enhance accessibility for both researchers and automated systems.

Recently, Large Language Models (LLMs) have shown great promise in handling a multitude of natural language tasks, including tackling mathematical reasoning problems \cite{ahn2024large,yue2023mammoth,azerbayev2024llemmaopenlanguagemodel,deepseek-math}. Out of the common benchmark suite for evaluating LLMs, datasets such as GSM8k \cite{cobbe2021gsm8k} and MATH \cite{hendrycksmath2021} remained central in the development of reasoning models \cite{o12024openai,deepseek-r1}, and continue to be challenging even for the larger, proprietary models \cite{arora-etal-2023-llms}. 

Current mathematics benchmarks and datasets have focused solely on English, mostly disregarding other low-resourced languages. The tacit requirement for using AI tools is fluency in English \cite{shi2022language}. However, mathematical reasoning ability is independent of the underlying language \cite{sep-language-thought} and Anglo-centric models have been shown to exhibit the same biases of the English language, even when prompted in other languages \cite{wendler2024llamas,wang2023not,liu2023multilingual}. The focus on datasets and models in a language other than English allows the democratization of learning for underrepresented languages and cultures.  

Recently, Romanian LLM development has started to flourish with initiatives such as OpenLLM-Ro \cite{masala2024openllmro}, having fine-tuned several LLMs on Romanian text. However, for evaluation, the authors used translated versions of popular English datasets and several native Romanian benchmarks, but no evaluation is performed on dedicated reasoning tasks in Romanian. Aside from code generation \cite{cosma-etal-2024-rocode,dumitran2024competitive}, currently there is no reasoning benchmark for Romanian.

In this work, we propose \textbf{RoMath}\footnote{GitHub: \href{https://github.com/cosmaadrian/romath}{github.com/cosmaadrian/romath}}\footnote{Huggingface: \href{https://hf.co/datasets/cosmadrian/romath}{hf.co/datasets/cosmadrian/romath}}, a Romanian mathematical reasoning benchmarking suite comprised of three datasets, \textit{Baccalaureate}, \textit{Competitions} and \textit{Synthetic}, each with its own particularities. RoMath aims to provide a comprehensive benchmark suite, having high-school-level problems across multiple domains (linear and abstract algebra, calculus, limits, geometry, probabilities) and across multiple levels of difficulty, ranging from easy calculations, to baccalaureate-level problems, to more difficult, proof-centric, competition-level problems. The purpose of RoMath is to provide a mathematical benchmark for Romanian and to stimulate the development of enhanced reasoning capabilities of non-English LLMs.

This work makes the following contributions:
\begin{enumerate}
    \item We construct and release \textbf{RoMath}, a novel mathematical reasoning benchmark suite with 76,910 problem statements in Romanian, consisting of three subsets, each with its own particularities and difficulty levels:  \textit{Baccalaureate} (5,777 problems), \textit{Competitions} (1,133 problems) and \textit{Synthetic} (63,000 problems). We collect and curate math problems using a semi-automatic workflow using foundational LLMs for providing structured output from unstructured raw OCR input and annotating problems with relevant metadata.
    
    \item We provide a comprehensive benchmark of several English and Romanian open-weight LLMs under several common scenarios - zero-shot, LoRA fine-tuning \cite{hu2022lora} and training with verifiable rewards using GRPO \cite{deepseek-math}. Furthermore, we provide an evaluation procedure using an LLM-as-a-judge paradigm \cite{zheng2023judgingllmasajudgemtbenchchatbot} for proofs, and analyze its performance to properly estimate solution correctness. 
    
    \item We show that simple translation of problem statements is not enough, as sub-par translations of precise mathematical language significantly reduces performance. Consequently, we emphasize the need for more dedicated resources in languages other than English.
\end{enumerate}

\section{Related Work}
\begin{table*}[!hbt]
    \centering
    \resizebox{0.75\linewidth}{!}{
        \begin{tabular}{llccp{7cm}}
            \textbf{Name} & \textbf{\# Problems} & \textbf{Level} & \textbf{Language} & \textbf{Notes}\\
            \toprule
            APE210K \cite{zhao2020ape210k}& 210K & \elementary & Chinese &  Requires basic arithmetic and common sense \\
            MATH23K \cite{ling2017program}& 23K & \elementary & Chinese & Contains questions, answers and rationales \\
            CMATH \cite{wei2023cmath}& 1.7K & \elementary & Chinese & Contains number of reasoning steps \\
            ARMATH \cite{alghamdi-etal-2022-armath}& 6K & \elementary & Arabic & Novel problems and inspired by MATH23K \\
            GSM8K \cite{cobbe2021gsm8k} & 8.5k  & \middleschool & English &Linguistically diverse. \\ 
            MATH \cite{hendrycksmath2021} & 12.5k & \highschool & English & Problems are put into difficulty levels 1-5 \\
            PRM800K \cite{lightman2023let} & 12k & \highschool & English & MATH w/ step-wise labels \\
            MathOdyssey \cite{fang2024mathodyssey} & 387 & \highschool \college & English & Problems from GAIC Math 2024 \\
            ARB \cite{sawada2023arb} & 105 & \college & English & Contest problems and university math proof \\ 
            AQUA \cite{ling2017program} & 100K & \college & English & GRE\&GMAT questions \\
            \midrule
            \textbf{RoMath} \\
            \hspace{3mm}\emojiro \hspace{1mm} \textbf{\textit{Baccalaureate}} & 5.8k & \highschool & Romanian & Easy-medium, baccalaureate problems.\\
            \hspace{3mm}\emojiro \hspace{1mm} \textbf{\textit{Competitions}} & 1.1k & \middleschool\highschool & Romanian & Difficult, proofs, competition problems.\\
            \hspace{3mm}\emojiro \hspace{1mm} \textbf{\textit{Synthetic}}  & 63k & \highschool & Romanian & Calculation-based, procedurally generated.\\
        \end{tabular}
    }
    \captionsetup{justification=centering}
    \caption{Comparison with other mathematical reasoning benchmarks. RoMath is the only Romanian mathematics benchmark outside of translated versions of English benchmarks. Table adapted from \citet{ahn2024large}. \newline \elementary = Elementary, \middleschool = Middle School, \highschool = High School, \college = College.}
    \label{tab:dataset-comparison}
\end{table*}

\textbf{Pretraining datasets for mathematics.} Interest in representation learning of mathematical expressions and text has existed in the past \cite{peng2021mathbert,collard2022extracting}. However, beyond representation learning, with the recent success of LLMs in a wide range of tasks, there has been increased attention to training and evaluating mathematical reasoning of LLMs. For pretraining, the general approach is to filter Common Crawl web pages and PDFs to obtain high quality math tokens. For instance, datasets such as MathWebPages \cite{lewkowycz2022solving}, ProofPile \cite{azerbayev2023proofnet} and OpenWebMath \cite{paster2023openwebmath} are used to pretrain high performing LLMs specialized in math such as Minerva \cite{lewkowycz2022solving} and LLema \cite{azerbayev2023llemma}.

\noindent\textbf{Mathematical reasoning benchmarks.} Regarding benchmarks, the most popular dataset is GSM8K \cite{cobbe2021gsm8k}, containing middle-school Math Word Problems (MWPs). An improved variant that contains process supervision (i.e., supervision at each intermediary reasoning step) is PRM800K \cite{lightman2023let}. However, these datasets are regarded as too simple to demonstrate advanced mathematical reasoning of LLMs. Consequently, MATH \cite{hendrycksmath2021} is a comparatively more difficult dataset, containing high-school problems from domains such as calculus, linear algebra, geometry and number theory. MathVISTA \cite{lu2024mathvista} is another similar benchmark, that contains mathematical reasoning in visual contexts (e.g., plots, natural images, functions).

Aside from simple word problems \cite{cobbe2021gsm8k} and datasets focused on QA-type problems, more difficult competition-level benchmarks have been proposed. For instance, ARB \cite{sawada2023arb} is a dataset comprised of problems from math competitions and problems from specialized books, with special care taken to avoid data contamination. While it contains problems that require proofs, ARB only contains 105 problems. MathOdyssey \cite{fang2024mathodyssey} contains difficult high-school and university-level problems but it is similarly small, as it contains only 387 problems.

\noindent\textbf{Non-English benchmarks.} Regarding datasets in languages other than English, there have been efforts in Arabic with datasets such as ArMATH \cite{alghamdi-etal-2022-armath} and Chinese with Ape210k \cite{zhao2020ape210k},  Math23k \cite{ling2017program}, CMath \cite{wei2023cmath}. Otherwise, outside of (automatically) translated versions of popular sets such as GSM8k \cite{masala2024openllmro}, as far as we know, no datasets currently exist for Romanian or other Latin languages. 

\noindent\textbf{Comparison with prior work.} Table \ref{tab:dataset-comparison} shows a comparison between similar datasets and RoMath. RoMath comprises middle-school, high-school and competitive high-school problems in Romanian covering multiple subjects and types of problems (proofs, calculations, equations, etc.). Different from prior datasets, RoMath is the first dedicated resource for mathematical reasoning in Romanian, a low-resource language of $\sim$23M speakers, which has its unique linguistic particularities \cite{dinu2005syllabic,dinu2007clustering}.

\section{Method}
We describe below the process for collecting \textit{Baccalaureate} and \textit{Competitions}, the two subsets that are collected by crawling publicly available PDFs. The \textit{Synthetic} subset is comprised of programmatically generated problems directly in Romanian.

\begin{figure*}[hbt!]
    \centering
    \includegraphics[width=0.75\linewidth]{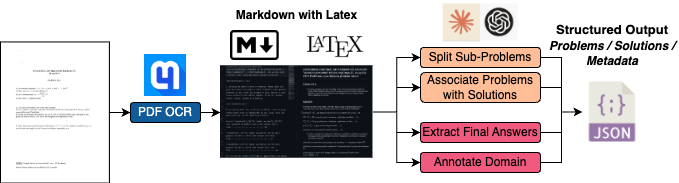}
    \caption{Overall diagram of our approach to curating problems from existing PDFs. We employ MathPix \cite{mathpix} to OCR PDFs and obtain markdown with LaTeX formatting for mathematical statements. We further process the markdown using proprietary LLMs to split into sub-problems, associate problems with the appropriate solution and annotate each problem with metadata.}
    \label{fig:construction}
\end{figure*}

\subsection{Dataset Construction}
In order to construct a high quality set of mathematical problems paired with solutions, we crawl publicly available PDFs from country-wide mathematics competitions and questions from the Romanian baccalaureate exam. Figure \ref{fig:construction} showcases our approach. After collecting raw PDFs, usually having separate documents for problem sets and their respective solutions, we utilize an academic document-focused OCR (i.e., MathPix \cite{mathpix}) to extract the underlying text and mathematical formulas / statements in LaTeX format. The final output is represented in Markdown format. 

To parse the content, instead of relying on brittle handcrafted rules and regex expressions, we utilize a commercial LLM (i.e., Claude 3 Sonnet \cite{anthropic2024claude}) to parse the raw text and to output structured JSON from unstructured Markdown. The LLM is provided with several examples of how to structure the final JSON (see Appendix \ref{sec:appendix}, Table \ref{tab:prompt-crawl} for the system prompt). The JSON output contains the LaTeX-formatted problem statement and its appropriate solution. Finally, we again utilize a commercial LLM to annotate the domain of the problem and to extract final answers for non-proof problems for easier evaluation (similar to \citet{hendrycksmath2021}, we enclose the final answer, if it exists, into a \texttt{$\backslash$boxed\{\}} tag). If a problem contains multiple sub-problems, we ensure that each sub-problem is self-contained and that the solution does not rely heavily on previous sub-problems' solutions. To split a problem into sub-problems, we used a prompt (presented in Appendix \ref{sec:appendix}, Table \ref{tab:prompt-crawl}) with specific instructions for parsing the data and output sub-questions that are self-contained. For example, if a problem is structured as follows:

{
\centering
\noindent \textit{<problem\_statement>}\\
\noindent \textit{<question\_1>}\\
\noindent \textit{<question\_2>}\\
}
The output is formatted as two separate standalone problems:

{\centering
\noindent \textit{<problem\_statement> <question\_1>}\\
\noindent \textit{<problem\_statement> <question\_2>}\\
}

\begin{figure}[t!]
    \centering
    \includegraphics[width=1\linewidth]{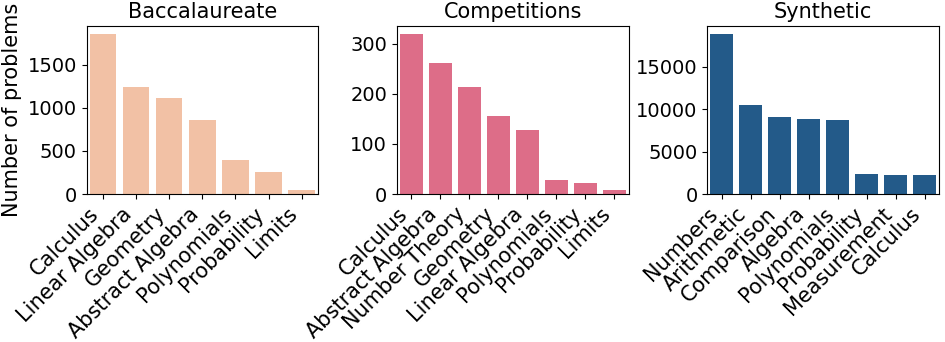}
    \caption{Distribution of the number of problems per domain for \textit{Baccalaureate}, \textit{Competitions} and \textit{Synthetic}.}
    \label{fig:datasets-distribution}
\end{figure}

Additionally, through manual inspection, we further removed any sub-questions that contained references to previous sub-questions (e.g. “Using the result from a) compute […]”). Figure \ref{fig:datasets-distribution} shows the distribution of problems per domain.

\subsection{RoMath Suite}

RoMath is comprised of three subsets: \textit{Baccalaureate}, \textit{Competitions} and \textit{Synthetic}. By its construction, each subset of RoMath features problems that require both single-step and multi-step reasoning for correctly solving problems. Usually, single-step reasoning problems involve simple calculations, while multi-step reasoning problems require solving intermediate solutions to reach a valid conclusion. Table \ref{tab:romath-examples} showcases selected examples from each subset.

\begin{table}
    \begin{tcolorbox}[
        enhanced,
        attach boxed title to top left={xshift=6mm,yshift=-3mm},
        colback=white,
        colframe=gray!20,
        colbacktitle=white,
        fonttitle=\bfseries\color{black},
        boxed title style={size=small,colframe=gray!20,rounded corners},
        rounded corners,
        ]
        \vspace{2mm}
        \tiny
        \textbf{\underline{Synthetic}} \\
        \textbf{\textit{Problem Statement}:} Care este rezultatul împărțirii lui -54 la -36495?\\
        \textbf{\textit{Solution}:} $\frac{6}{4055}$ \\
        \textbf{\textit{Problem Statement}:} Fie $u$ definit ca fiind $(2 - \frac{18}{15})\cdot5$. Găsește valoarea lui $r$ din ecuațiile $u \cdot y + 8 = 0$, $-28 = 3 \cdot r + 4 \cdot y - 5$.\\
        \textbf{\textit{Solution}:} $-5$ \\
        \textbf{\underline{Baccalaureate}} \\
        \textbf{\textit{Problem Statement}:} Se consideră funcția \(f: \mathbf{R} \rightarrow \mathbf{R}, f(x)=e^{x}-x\). Să se calculeze \(\int_{0}^{1} f(x) d x\). \\
        \textbf{\textit{Solution}:} \(\int_{0}^{1} f(x) d x=e-\frac{3}{2}\). Soluția finală este $\boxed{e-\frac{3}{2}}$ \\
        \textbf{\textit{Problem Statement}:} Se consideră funcțiile \(f_{n}: \mathbf{R} \rightarrow \mathbf{R}, f_{1}(x)=x^{3}-3 x^{2}+3 x\) şi \(f_{n+1}(x)=\left(f_{1} \circ f_{n}\right)(x)\), \(\forall n \in \mathbf{N}^{*}, \forall x \in \mathbf{R}\). Să se rezolve în mulțimea numerelor reale ecuația \(f_{1}(x)+f_{2}(x)+f_{3}(x)-3=0\).\\
        \textbf{\textit{Solution}:} Observăm că \(x=1\) este soluție. Dacă \(x>1 \Rightarrow f_{1}(x)+f_{2}(x)+f_{3}(x)>3\). Analog dacă \(x<1 \Rightarrow f_{1}(x)+f_{2}(x)+f_{3}(x)<3\). Deci \(x=1\) este soluție unică. Soluția finală este $\boxed{x = 1}$\\
        \textbf{\underline{Competitions}} \\
        \textbf{\textit{Problem Statement}:} Se consideră numerele complexe \(u, v\) și \(z\) astfel încât \(|u|=|v|=1\) și \(|u+v|=\sqrt{3}\). Să se demonstreze că: \(\mathrm{u} \cdot \overline{\mathrm{v}}+\overline{\mathrm{u}} \cdot \mathrm{v}=1\). \\
        \textbf{\textit{Solution}:} \(|\mathrm{u}+\mathrm{v}|^{2}=3 \Leftrightarrow(\mathrm{u}+\mathrm{v})(\overline{\mathrm{u}}+\overline{\mathrm{v}})=3 \Leftrightarrow \mathrm{u} \cdot \overline{\mathrm{v}}+\overline{\mathrm{u}} \cdot \mathrm{v}=1\) \\
        \textbf{\textit{Problem Statement}:} Să se rezolve, în \(\mathbb{R}\), inecuația: \(\left(\frac{2}{5}\right)^{\frac{6-5 x}{5 x+2}} \leq \frac{25}{4}\) \\
        \textbf{\textit{Solution}:} \(\left(\frac{2}{5}\right)^{\frac{6-5 x}{5 x+2}} \leq \frac{25}{4}=\left(\frac{5}{2}\right)^{-2} \Rightarrow \frac{6-5 x}{5 x+2} \geq-2\), cu \(x \neq-\frac{2}{5}\). \(\frac{6-5 x}{5 x+2} \geq-2 \Rightarrow \frac{x+2}{5 x+2} \geq 0 \Rightarrow x \in(-\infty,-2] \cup\left(-\frac{2}{5}, \infty\right)\) \\
    \end{tcolorbox}
    \caption{Qualitative examples from each subset of RoMath.}
    \label{tab:romath-examples}
\end{table}

\textbf{\textit{Baccalaureate}} is composed of problems and solutions from the Romanian Baccalaureate exam. The Romanian Baccalaureate is a country-wide exam for graduating high-school students, comprised of three subjects, each with several problems and sub-problems. Students taking the Baccalaureate exam consider the calculus problems, such as solving an integral or computing a limit, to be the most difficult. However, the calculus problems rarely require more than 2 steps of reasoning and some calculation. This subset contains a total of 5777 problems: 4.3k problems for training and 1.48k testing. Most problems (4617 / $\sim$ 80\%) in this subset are verifiable (i.e., have a single final answer), while some (1160 / $\sim$ 20\%) require proofs. Furthermore, 4038 / $\sim$ 69\% problems in this category also have intermediate steps provided in the ground-truth solution. In this set, there are multiple domains, with varying difficulty: geometry, combinatorics, abstract algebra, linear algebra, calculus (integrals and derivatives), and limits. In all categories, we discarded any problem that required reasoning over images or plots. For instance, geometry problems do not have an accompanying drawing or figure. If we encountered images in the source PDFs, we removed the problem entirely through manual inspection. The \textit{Baccalaureate} subset includes only standalone geometry problems: an example of such a problem would be the following (here, translated in English for convenience): \textit{“In a Cartesian coordinate system \(x O y\) we consider the points \(A_{n}(n, 0)\) and \(B_{n}(0, n)\), with \(n \in\{1 ,2,3\}\). Calculate the area of the triangle \(A_{1} A_{2} B_{2}\).” }

\textbf{\textit{Competitions}} is the hardest subset of RoMath, containing 1133 problems sourced from mathematics competitions, with problems ranging from local to inter-county and olympiad events, out of which 804 problems are for training and 329 for testing. Different from \textit{Baccalaureate}, this subset also contains middle-school problems. Around half of the problems (594 / $\sim$ 52\%) require proofs for a complete solution, while the rest are directly verifiable. Almost all problems in this subset have intermediate explanations. The problems in \textit{Competitions} are considered hard, requiring insight and problem-solving skills outside of simple symbol manipulations \cite{polya1971solve}. The extraction and post-processing steps are identical to those in \textit{Baccalaureate}.

\textbf{\textit{Synthetic}} is programmatically generated, using the approach of \citet{saxton2019analysing}, in which we manually translate the source key-phrases and formulations in Romanian. Problems in this subset have a single final answer. Problems are mostly algebraic in nature, and are split into arithmetic, calculus, derivatives, integrations, polynomials, composition of problems, comparisons, manipulating expressions (e.g., simplification), numbers, measurements. All problems in this subset are verifiable, having only a single final answer provided, without intermediate steps, making it difficult to directly provide an answer without the use of external tools or chain-of-thought prompting. In contrast to the other sub-sets in RoMath, there is less linguistic variation present in problem statements, but there is complete control over correctness and difficulty. We emphasize that \textit{Synthetic} is \textit{not} a direct translation of the problems contained in DeepMind Mathematics \cite{saxton2019analysing}, but rather a manual translation of the phrases that are used to generate the problems. As such, one could generate an indefinite number of problems. We make the code for generating \textit{Synthetic} open-source and provide, for convenience, 63k generated problems, out of which 55.9k problems for training and 7.1k for testing.

\subsection{Evaluation Procedure}
Generally, there are two ways to evaluate solutions: \textit{(i)} for verifiable problems (i.e., containing a single final answer), correctness is estimated by direct string comparison between the model answer and the correct answer after normalization \cite{hendrycksmath2021,cobbe2021gsm8k} and \textit{(ii)} using a proof-checker for problems requiring proofs \cite{li2024survey}. 

Evaluating the correctness of a solution to a mathematics problem for proof problems is still an open problem. Using a proof-checker is not always feasible as it requires the problems and solutions to already be formalized into the language of the proof-checker \cite{Trinh2024alphageometry}, an unrealistic requirement for most mathematics written in natural language. For proof-type problems, where it is necessary to check for correctness at every reasoning step in natural language, there is no consensus on the evaluation procedure outside of formal proof-checkers. 

However, more recent methods \cite{fang2024mathodyssey} have adopted a "soft" evaluation of proof solutions by employing an external judge LLM tasked to output a correctness score given the problem statement, the correct solution and a provided solution to be scored. 

To evaluate solutions to RoMath, we propose the following procedure: For evaluating verifiable problems, we adopt the procedure from \cite{hendrycksmath2021} for string comparison after the solutions are normalized; this requires the model to output solutions in a \texttt{$\backslash$boxed\{\}} tag. However, if the model does not provide the solution in this format or if the problem requires a proof, we employ a judge LLM to estimate correctness, inspired by several other works \cite{zheng2023judgingllmasajudgemtbenchchatbot,fang2024mathodyssey}. Since the use of proprietary LLMs is prohibitively expensive, there are concerns with reproducibility, and there is no information on the architecture and training dataset, we use existing open-weight models.

\section{Baselines and Results}
\subsection{Judge Evaluation}
Very few analyses have been performed to gauge the performance of the judge LLM: for instance, a more recent study \cite{bavaresco2024llms} showed that LLMs exhibit a large variance across datasets in correlation to human judgments. However, there is no study estimating the performance of judge LLMs for mathematical reasoning in a language other than English. Using LLMs as judges is a reasonable proxy for estimating performance, and we show in Section \ref{sec:judge-ablation} that performance is relatively robust across multiple judges.

In this section, we conduct an analysis of the performance of multiple open-weight judge models in evaluating solution correctness in Romanian, using both Romanian and English system prompts (see Appendix \ref{sec:appendix} Tables \ref{tab:romanian-judge-prompt} and \ref{tab:english-judge-prompt}). 

We programmatically construct a dataset of 300 problems from the training sets of \textit{Baccalaureate} and \textit{Competitions} containing correct and incorrect solutions. Correct solutions are constructed by symbol changes \cite{meadows2023symbolic} and removal of natural language text (keeping only mathematical expressions) of the original ground-truth solution, and incorrect solutions are either original solutions with some operators / number modified (e.g., $+$ sign changed to $-$, or $<$ symbol changed to $\geq$, and others) or a similar solution, but not exactly the same, from another problem based on the Levenstein distance. 

In Table \ref{tab:judge-performance}, we showcase the performance of multiple LLMs-as-judges on our programmatically generated dataset to estimate judge performance. We tested Qwen2 \cite{qwen2} family of models, as well as the math-specialized variant Qwen2-Math-7B, deepseek-math \cite{deepseek-math}, Phi-3 \cite{abdin2024phi3technicalreporthighly}, Llama3-70B \cite{dubey2024llama}, Mathstral \cite{mathstral}, and Mixtral-8x7b \cite{jiang2024mixtralexperts}. For this synthetical dataset, we obtained that Qwen2-7B-Instruct prompted in English obtained the best overall results of 91\% accuracy judging solution correctness. Surprisingly, the math-specialized models severely underperformed at this task. As such, unless otherwise specified, we used Qwen2-7B-Instruct prompted in English as a judge for the rest of the non-verifiable results.

\begin{table}[hbt!]
    \centering
    \resizebox{\linewidth}{!}{
    \begin{tabular}{lp{1.2cm}|rrr}
        \textbf{Judge Model} & \textbf{System Prompt} &   \textbf{Acc.} $\uparrow$ &  \textbf{FPR} $\downarrow$ &  \textbf{FNR} $\downarrow$\\
        \midrule
deepseek-math-7b-instruct &     ro &      0.51 & 0.66 & 0.31 \\
 Meta-Llama-3-70B-Instruct &     ro &      0.86 & 0.24 & 0.03 \\
 Mixtral-8x7B-Instruct-v0.1 &     ro &      0.84 & 0.27 & 0.03 \\
          Qwen2-Math-7B-Instruct &     ro &      0.87 & 0.22 & 0.03 \\
               Qwen2-7B-Instruct &     ro &      \textbf{0.90} & \textbf{0.17} & 0.02 \\
        \midrule
deepseek-math-7b-instruct &     en &      0.74 & 0.31 & 0.22 \\
 Meta-Llama-3-70B-Instruct &     en &      0.84 & 0.29 & \textbf{0.01} \\
 Mixtral-8x7B-Instruct-v0.1 &     en &      0.84 & 0.27 & 0.03 \\
          Qwen2-Math-7B-Instruct &     en &      0.89 & 0.16 & 0.06 \\
               Qwen2-7B-Instruct &     en &      \textbf{0.91} & 0.12 & 0.05 \\
    \end{tabular}}
    \caption{Judge LLM performance on a programmatically generated dataset of correct and incorrect student solutions.}
    \label{tab:judge-performance}
\end{table}

\subsection{Model Benchmark}

We chose to benchmark several open-weight LLMs, as opposed to proprietary models, to make the benchmark reproducible and to avoid unnecessary inference costs. We evaluated the performance under 0-shot and LoRA fine-tuned models for Qwen2-7B, Phi-3, Meta-Llama-8B and math-specialized variants such as Qwen2-Math-7B, deepseek-math-7b, Mathstral-7b. We evaluated larger models under 0-shot setting: Meta-Llama-70B and Mixtral-8x7B. Furthermore, we also evaluated Romanian-specialized models trained with continual pretraining on Romanian tokens, but with no focus on math tokens: RoLlama3-8B and RoMistral-7b \cite{masala2024openllmro}. In Appendix \ref{sec:appendix}, Table \ref{tab:romanian-prediction-prompt}.

For fine-tuning the models, we used LoRA \cite{hu2022lora}, using a rank of 8, alpha of 32 and dropout of 0.1, applied on all linear layers. Due to hardware limitations, we used a small batch size of 4 and a learning rate of 0.00002 with a linear decay over the 3 training epochs. 

\begin{table}[hbt!]
    \centering
    \resizebox{1.0\linewidth}{!}{
\begin{tabular}{cllrr|rr|rr}
\toprule
            & & & \multicolumn{2}{c}{\textbf{Baccalaureate}} &  \multicolumn{2}{c}{\textbf{Competitions}} &  \multicolumn{2}{c}{\textbf{Synthetic}} \\
            &\textbf{Model} &      \textbf{Scenario} &  \textbf{Accuracy} &  \textbf{F1} &  \textbf{Accuracy} &  \textbf{F1} &   \textbf{Accuracy} &  \textbf{F1} \\
\midrule
       \multirow{4}{*}{\rotatebox{90}{\textbf{Romanian}}} & OpenLLM-Ro/RoLlama3-8b-Instruct &     0-shot &        0.50 &  0.67 &        0.48 &  0.65 &      0.18 & 0.30 \\
       & & fine-tuned &        0.18 &  0.31 &        0.50 &  0.67 &      -- & -- \\
      \cmidrule{3-9}
      & OpenLLM-Ro/RoMistral-7b-Instruct &     0-shot &        0.50 &  0.66 &        0.44 &  0.61 &      0.16 & 0.27 \\
      & & fine-tuned &        0.18 &  0.31 &        0.36 &  0.53 &      -- & -- \\
      \midrule
                \multirow{6}{*}{\rotatebox{90}{\textbf{General-Purpose}}} & Qwen/Qwen2-7B-Instruct &     0-shot &        0.40 &  0.57 &        0.55 &  0.71 &      0.29 & 0.45 \\
                & & fine-tuned &        0.54 &  0.70 &        0.48 &  0.65 &      -- & -- \\
 \cmidrule{3-9}
      & microsoft/Phi-3-mini-4k-instruct &     0-shot &        0.36 &  0.53 &        0.33 &  0.50 &      0.07 & 0.14 \\
      && fine-tuned &        0.25 &  0.40 &        0.41 &  0.58 &      -- & -- \\
     \cmidrule{3-9}
   & meta-llama/Meta-Llama-3-8B-Instruct &     0-shot &        0.34 &  0.51 &        0.53 &  0.69 &      0.25 & 0.40 \\
   && fine-tuned &        0.18 &  0.31 &        0.33 &  0.49 &      -- & -- \\

 \midrule
           \multirow{6}{*}{\rotatebox{90}{\textbf{Math-Specialized}}} & Qwen/Qwen2-Math-7B-Instruct &     0-shot &        0.32 &  0.48 &        0.55 &  0.71 &      0.27 & 0.43 \\
           && fine-tuned &        0.48 &  0.65 &        0.57 &  0.73 &      -- & -- \\
          \cmidrule{3-9}
 &deepseek-ai/deepseek-math-7b-instruct&     0-shot &        0.56 &  0.72 &        0.59 &  0.74 &      0.21 & 0.35 \\
 && fine-tuned &        0.29 &  0.44 &        0.56 &  0.72 &      -- & -- \\
\cmidrule{3-9}
           & mistralai/Mathstral-7b-v0.1 &     0-shot &        0.30 &  0.46 &        0.61 &  0.75 &      0.36 & 0.53 \\
           && fine-tuned &        0.21 &  0.34 &        0.56 &  0.71 &      -- & -- \\

\midrule
 \multirow{4}{*}{\rotatebox{90}{\textbf{Large}}}& meta-llama/Meta-Llama-3-70B-Instruct &     0-shot &        0.25 &  0.40 &        0.22 &  0.36 &      0.10 & 0.19 \\
  &&     5-shot &        0.08 &  0.15 &        0.09 &  0.16 &      0.07 & 0.13 \\
         \cmidrule{3-9}
 &mistralai/Mixtral-8x7B-Instruct-v0.1 &     0-shot &        0.43 &  0.60 &        0.60 &  0.75 &      0.32 & 0.48 \\
  &&     5-shot &        0.25 &  0.40 &        0.25 &  0.40 &      0.24 & 0.38 \\
\bottomrule
\end{tabular}
}
    \caption{Results for various open-weight LLMs on \textit{Baccalaureate}, \textit{Competitions} and \textit{Synthetic}, under 0-shot and fine-tuned scenarios.}
    \label{tab:romath-sota}
\end{table}

\begin{figure}[hbt!]
    \centering
    \includesvg[width=1.0\linewidth]{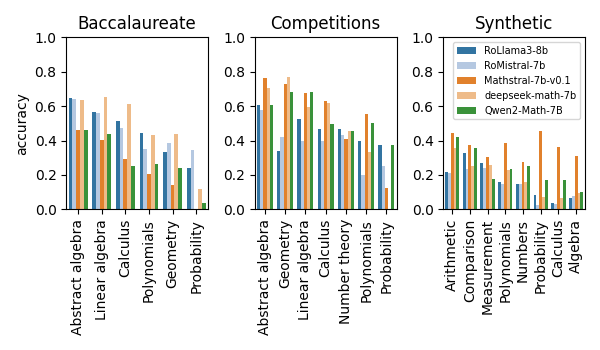}
    \caption{Performance of Romanian models and math-specialized models on each domain from each RoMath subset.}
    \label{fig:perf-per-domain}
\end{figure}

In Table \ref{tab:romath-sota}, we showcase the performance of the models under zero-shot, and LoRA-fine-tuned scenarios. The best performing model on the \textit{Baccalaureate} subset is deepseek-math-7b, while on \textit{Competitions} and \textit{Synthetic} Mathstral-7b obtains the best results. However, the Romanian models, RoLlama-8b and RoMistral-7b obtain competitive results on all subsets, which can be attributed to their better understanding of Romanian text compared to English-focused models, since specialization on mathematical text did not receive a particular emphasis during training. Surprisingly, we obtained that fine-tuning does not always result in improved performance. Fine-tuning improves performance on \textit{Baccalaureate} for Qwen2-7b and Qwen2-Math-7b, while on \textit{Competitions}, RoLlama-7b, Phi-3, Qwen2-Math-7b benefit from further fine-tuning. One possible explanation is that the solutions present in RoMath are qualitatively different (different formatting, explanation style) than solutions present in other math datasets \cite{cobbe2021gsm8k,hendrycksmath2021} and Chain-of-Thought style prompting \cite{wei2022chain}. Further investigation on this effect is left as future work. In Figure \ref{fig:perf-per-domain}, we show extended results per problem domain for each dataset. Qualitative examples of generated solutions are shown in the Appendix \ref{sec:appendix} Tables \ref{tab:correct-predictions} and \ref{tab:incorrect-predictions}. 

\subsection{Training with Verifiable Rewards}

Since a significant proportion of problems in RoMath includes intermediate steps and are verifiable, we tested if the problems are of sufficiently high quality to enable training with rewards. We adopt a part of the training procedure from \citet{deepseek-math}, and fine-tune two variants of the Llama3.2 \cite{dubey2024llama} (1B and 3B parameters) and Qwen2 \cite{qwen2} (0.5B and 1.5B) family of models. For supervised fine-tuning (SFT), we train on all problems from \textit{Baccalaureate} and \textit{Competitions} that contain intermediate steps to force the model to conform to the specified output format of \textit{<raționament> [\dots] </raționament> <răspuns> [\dots] </răspuns>}. 

Further, we train using GRPO \cite{deepseek-math} with 4 completions per prompt on all verifiable problems from \textit{Baccalaureate} and \textit{Competitions}, using only a correctness reward and a format reward. Figure \ref{fig:grpo} shows the performance on the verifiable problems from the \textit{Baccalaureate} subset for this setting. Training with rewards reliably boosts performance compared to only supervised fine-tuning. As such, RoMath can be a useful resource for training Romanian reasoning models.

\begin{figure}[hbt!]
    \centering
    \includesvg[width=0.8\linewidth]{images/verifiable.svg}
    \caption{Performance of GRPO-trained LLama-3.2 and Qwen2 on on a subset of \textit{Baccalaureate} that has verifiable answers.}
    \label{fig:grpo}
\end{figure}

\subsection{Impact of the Judge Model}
\label{sec:judge-ablation}
\begin{figure}[hbt!]
    \centering
    \includesvg[width=1.0\linewidth]{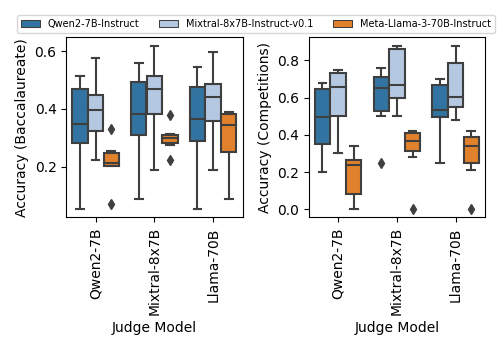}
    \caption{Performance using different judge models.}
    \label{fig:perf-per-judge}
\end{figure}

In Figure \ref{fig:perf-per-judge}, we compared multiple judge models to gauge their effect on downstream performance. Based on Table \ref{tab:judge-performance}, we used Qwen2-7B, Llama-70B and Mixtral-8x7b as judges and used them to evaluate the performance of the same Qwen2-7B, Llama-70B and Mixtral-8x7b. We chose the same judges and downstream models to check if judges prefer the output of their own model. From Figure \ref{fig:perf-per-judge} we find that judges do not have "favorites". However, we do find that, for example, in \textit{Competitions}, where there are more proofs than in \textit{Baccalaureate}, the Llama-70B and Mixtral-8x7b judges give higher scores on average, which might explain why results on the \textit{Competitions} subset are higher: judges might artificially inflate results. While the differences between judges are small, there is a clear ascending trend between them.

\subsection{Translating Romanian Problems to English}

Translating domain-specific technical language is non-trivial. \citet{al2024bridging} identified multiple linguistic challenges that make translation difficult. Translating mathematics is challenging due to the need for precise language, as even slight ambiguities can alter meaning. Although mathematical concepts are universal, their interpretation varies across cultures. Additionally, mathematical symbols and notations are not always standardized across languages, and mathematical terms lack direct equivalents in other languages leading to potential confusion if not properly accounted for. 

We used the NLLB \cite{nllb2022} family of models (600M, 1.3B, and 3.3B) to translate from Romanian to English the test sets for \textit{Baccalaureate} and \textit{Competitions}, as the models have established numerical benchmarks on Romanian to English translation. Directly translating the full problem statement and solution resulted in "gibberish" translations due to the mathematical symbols present in the text.  As such, we opted to keep the LaTeX-delimited section intact and only translate the surrounding natural language. While this approach might lose some of the larger context, we found it to be the only satisfactory approach. Still, the resulting translations contain unnatural English formulations and sometimes spurious text. For instance, the problem statement \textit{"Se consideră funcția \(f: \mathbf{R} \rightarrow \mathbf{R}, f(x)=e^{x}-x+1\). Să se calculeze $\lim_{x \rightarrow 0} \frac{f(x)-f(0)}{x}$"} is translated as \textit{"It's considered function \(f: \mathbf{R} \rightarrow \mathbf{R}, f(x)=e^{x}-x+1\). Let's figure it out. $\lim_{x \rightarrow 0} \frac{f(x)-f(0)}{x}$ [♪ I'm not gonna let you down ♪]"}, in which the part "\textit{[♪ I'm not gonna let you down ♪]}" is introduced spuriously by the translation model. 

In Table \ref{tab:translation-sota}, we showcase the performance of math-specialized LLMs on the English-translated version of \textit{Baccalaureate} and \textit{Competitions} using the different sizes of NLLB. Compared to the original Romanian text, translating severely degrades performance. We found that performance improves with the translation model size, but up to a certain point. The main point of failure is handling the math LaTeX tokens without disrupting the surrounding text. The use of an LLM for translation might be more appropriate only if their reliability and control of their output are properly established, and proper benchmarks for translation in Romanian are in place.

\begin{table}[hbt!]
    \centering
    \resizebox{1.0\linewidth}{!}{
\begin{tabular}{cllp{2cm}p{2cm}l}
\toprule
          & \textbf{Model} &  \textbf{Translation Model}  &\textbf{Romanian Accuracy}&  \textbf{English-Translated Accuracy}  &\textbf{Diff.}\\
\midrule
           \multirow{9}{*}{\rotatebox{90}{\textbf{Baccalaureate}}} & & nllb-200-distilled-600M  & &      0.04  &{\color{red}\textbf{--0.28}}\\
           & Qwen/Qwen2-Math-7B-Instruct &           nllb-200-1.3B  &0.32&      0.03  & {\color{red}\textbf{--0.29}}\\
           & &           nllb-200-3.3B &  &      0.03  &{\color{red}\textbf{--0.29}}\\
          \cmidrule{3-6}
&  &  nllb-200-distilled-600M  &&      0.09  &{\color{red}\textbf{--0.47}}\\
& deepseek-ai/deepseek-math-7b-instruct &           nllb-200-1.3B  &0.56&      0.05  &{\color{red}\textbf{--0.51}}\\
&  &           nllb-200-3.3B  &&      0.07  &{\color{red}\textbf{--0.49}}\\
\cmidrule{3-6}
          && nllb-200-distilled-600M  &&      0.07  &{\color{red}\textbf{--0.23}}\\
          &mistralai/Mathstral-7b-v0.1 &            nllb-200-1.3B  &0.30&      0.07  &{\color{red}\textbf{--0.23}}\\
          & &           nllb-200-3.3B  &&      0.07  &{\color{red}\textbf{--0.23}}\\
\midrule
          \multirow{9}{*}{\rotatebox{90}{\textbf{Competitions}}} &  &  nllb-200-distilled-600M  &&      0.17  &{\color{red}\textbf{--0.38}}\\
          &Qwen/Qwen2-Math-7B-Instruct &           nllb-200-1.3B  &0.55&      0.20  &{\color{red}\textbf{--0.35}}\\
          & &            nllb-200-3.3B  &&      0.19  &{\color{red}\textbf{--0.36}}\\
          \cmidrule{3-6}
& &  nllb-200-distilled-600M  &&      0.09  &{\color{red}\textbf{--0.50}}\\
&deepseek-ai/deepseek-math-7b-instruct &           nllb-200-1.3B  &0.59&      0.12  &{\color{red}\textbf{--0.47}}\\
& &           nllb-200-3.3B  &&      0.10  &{\color{red}\textbf{--0.49}}\\
\cmidrule{3-6}
          & &  nllb-200-distilled-600M  &&      0.19  &{\color{red}\textbf{--0.42}}\\
          &mistralai/Mathstral-7b-v0.1 &            nllb-200-1.3B  &0.61&      0.21  &{\color{red}\textbf{--0.40}}\\
       & &           nllb-200-3.3B  &&      0.20  &{\color{red}\textbf{--0.41}}\\
\bottomrule
\end{tabular}
}
    \caption{Results on RoMath-Baccalaureate and RoMath-Competitions for math-specific LLMs in 0-shot setting with English-translated problems. Performance drops significantly due to poor quality translations.}
    \label{tab:translation-sota}
\end{table}

\section{Conclusions and Future Directions}

In this paper, we proposed RoMath, a benchmarking suite consisting of three datasets with mathematical problems written in Romanian: \textit{Baccalaureate}, \textit{Competitions} and \textit{Synthetic}. We detailed the construction process and composition for each subset and benchmarked several open-weight LLMs under different training and evaluation scenarios. We are the first to provide quantitative results for mathematical reasoning in Romanian.

Surprisingly, we found that mathematics problems written in Romanian can be properly handled by English-centric models, providing proper solutions in Romanian. It is unclear why this occurs, especially since such models are not explicitly trained on Romanian math tokens and most models have strong language filters to train only on English. Our results suggest that such LLMs would potentially receive a passing grade (i.e., more than 50\%) on the Romanian baccalaureate exam, scoring an average of $\sim$56\% across all problems in \textit{Baccalaureate}. 

An important future direction is reliable automatic annotations with chain-of-thought (CoT) traces for multilingual reasoning problems. Our results indicate that a significant factor in improving performance in mathematical reasoning is the presence of intermediate reasoning steps in the solutions. Performance is not reliably improved by fine-tuning without CoT, and the presence of more detailed solutions enables scalable training with reinforcement learning algorithms such as GRPO \cite{deepseek-math}. Currently, only a subset of RoMath contains intermediate steps for problem solutions, and further structured annotations could significantly increase the data quality.

\section*{Limitations}
\label{sec:lim}

The main limitation of this work is the use of an external LLM as a judge to estimate solution correctness, which might skew the results, artificially inflate performance. For example, some generated solutions for proof-type problems obtain the correct final result, but the intermediate steps are incorrect. In some cases, the judge model deemed these types of solutions as correct, whereas they are not. While this is an inherent limitation in literature for mathematics datasets that contain proofs, this is currently an open problem and there are on-going efforts to formalize proof verification \cite{alphaproof}. Furthermore, we argued that the proper way to evaluate solutions of generated proofs is by using an external proof verification tool such as Lean \cite{lean2015}.

\bibliography{refs}
\bibliographystyle{acl_natbib}

\newpage
\appendix

\section{Appendix}
\label{sec:appendix}

\begin{table}[hbt!]
    \begin{tcolorbox}[
        enhanced,
        attach boxed title to top left={xshift=6mm,yshift=-3mm},
        colback=white,
        colframe=gray!20,
        colbacktitle=white,
        fonttitle=\bfseries\color{black},
        boxed title style={size=small,colframe=gray!20,rounded corners},
        rounded corners,
        ]
        \vspace{2mm}
        \small

Given the following mathematics problems in Romanian formatted in MathPix markdown, make a JSON with subject and solution pairs, removing unnecessary boilerplate and extra problem identifiers. The JSON must contain the full problem definition and subject number (e.g. subject 1b). Each sub-question must contain the whole problem definition for completeness. Each subject must be self-contained. Do not output anything else besides the required JSON. Do not modify the latex describing the mathematical formulas.\\
Example (truncated):
"\\

PROBLEMS:\\

Se consideră matricea \(A=\left(\begin{array}{ll}1 & 1 \\ 1 & 0\end{array}\right)\) şi şirul \(\left(F_{n}\right)_{n \geq 0}\) definit prin relația de recurență \(F_{n+1}=F_{n}+F_{n-1}, n \in \mathbf{N}^{*}\), cu \(F_{0}=0, F_{1}=1\).\\

 a) Să se calculeze determinantul şi rangul matricei \(A\).\\\\

 b) Să se calculeze \(F_{2}\) şi \(F_{3}\).\\

SOLUTIONS:\\
a) \(\operatorname{det} A=-1 \neq 0 \Rightarrow \operatorname{rang} A=2 ;\) b) \(F_{2}=1, F_{3}=2\).
"\\
\\
Example Output JSON:
\\
{[}
  \{\\
    "subject": "1a",\\
    "definition": "Se consideră matricea \(A=\left(\begin{array}{ll}1 & 1 \\ 1 & 0\end{array}\right)\) şi şirul \(\left(F_{n}\right)_{n \geq 0}\) definit prin relația de recurență \(F_{n+1}=F_{n}+F_{n-1}, n \in \mathbf{N}^{*}\), cu \(F_{0}=0, F_{1}=1\). Să se calculeze determinantul şi rangul matricei \(A\).",\\
    "solution": "$\operatorname{det} A=-1 \neq 0 \Rightarrow \operatorname{rang} A=2$"\\
  \},\\
  \{\\
    "subject": "1b",\\
    "definition": "Se consideră matricea $A=\left(\begin{array}{ll}1 & 1 \\ 1 & 0\end{array}\right)$ şi şirul $\left(F_{n}\right)_{n \geq 0}$ definit prin relația de recurență $F_{n+1}=F_{n}+F_{n-1}, n \in \mathbf{N}^{*}$, cu $F_{0}=0, F_{1}=1$. Să se calculeze $F_{2}$ şi $F_{3}$.",\\
    "solution": "\(F_{2}=1, F_{3}=2\)"\\
  \}\\
{]}

In this example, each sub-problem is self-contained and is paired with the appropriate solution. The sub-problem identifiers (e.g., "a)" and "b)") are stripped. The latex markdown is left intact.
\\

Real Input:

    \end{tcolorbox}
    \caption{Claude 3 Sonnet prompt to format raw Markdown into structured JSON.}
    \label{tab:prompt-crawl}
\end{table}
\begin{table}[hbt!]
    \begin{tcolorbox}[
        enhanced,
        attach boxed title to top left={xshift=6mm,yshift=-3mm},
        colback=white,
        colframe=gray!20,
        colbacktitle=white,
        fonttitle=\bfseries\color{black},
        boxed title style={size=small,colframe=gray!20,rounded corners},
        rounded corners,
        ]
        \vspace{2mm}
        \small
        \{\\
            "role": "system",\\
            "content":
            """Ești un student olimpic la matematică care a participat și câștigat multiple concursuri internaționale de matematică. Rolul tău este să rezolvi probleme de matematică de liceu și să oferi soluții complete și corecte.
    Problemele care necesită demonstrații trebuie rezolvate complet cu toți pașii intermediari necesari. Problemele care au un singur raspuns final trebuie furnizate într-un format încadrat (`$\backslash$boxed{{}}`).
    Matematica trebuie scrisă în format LaTeX pentru a asigura claritatea și precizia soluțiilor. Textul in format LaTeX trebuie delimitat folosind simbolurile ` $\backslash$(` și `$\backslash$)`.
    Rezolvările incomplete sau incorecte vor fi evaluate cu scoruri mai mici. Asigură-te că răspunsurile sunt concise, fără prea multe explicații inutile.""",\\
        \},\\
        \# add few shot examples here: User => Problem statement, Assistant => Solution\\
        \{\\
            "role": "user",\\
            "content": """Care este rezolvarea următoarei probleme?\\
    
    \{problem\_statement\}""",\\
        \}
    \end{tcolorbox}
    \caption{Romanian prediction prompt.}
    \label{tab:romanian-prediction-prompt}
\end{table}
\begin{table*}[hbt!]
    \begin{tcolorbox}[
        enhanced,
        attach boxed title to top left={xshift=6mm,yshift=-3mm},
        colback=white,
        colframe=gray!20,
        colbacktitle=white,
        fonttitle=\bfseries\color{black},
        boxed title style={size=small,colframe=gray!20,rounded corners},
        rounded corners,
        ]
        \vspace{2mm}
        \small

    \{
        "role": "system",\\
        "content": """Asumă-ți rolul unui profesor de matematică responsabil cu evaluarea răspunsurilor studenților pentru o problemă de matematică în raport cu soluțiile corecte furnizate. Soluțiile pot include demonstrații, valori exacte, răspunsuri cu alegere multiplă sau aproximări numerice.\\

\#\# Criterii de Evaluare:\\
1. **Echivalență Matematică**: Evaluează răspunsurile pe baza echivalenței matematice, nu doar a preciziei numerice. Verifică dacă diferite expresii algebrice sau simbolice sunt echivalente. Asigură-te că sunt echivalențe precum $\backslash$( $\backslash$frac\{$\backslash$sqrt\{6\}-$\backslash$sqrt\{2\}\}\{2\} $\backslash$) fiind echivalent cu $\backslash$( $\backslash$sqrt\{2 - $\backslash$sqrt\{3\}\} $\backslash$).\\
2. **Scor**: Atribuie un scor de '1' pentru orice răspuns care se potrivește sau este echivalent cu soluția furnizată, fie că este o valoare exactă, o variantă de răspuns (de exemplu, A, B, C) sau o aproximare numerică corect rotunjită. Atribuie un scor de '0' pentru răspunsuri incorecte. Nu furniza niciun fel de explicație.\\
3. **Tratarea Alegerii Multiple**: Dacă soluția furnizată este o variantă de răspuns (de exemplu, A, B, C, D, E, F) și studentul identifică această alegere corect, trateaz-o ca fiind corectă. Dacă soluția este o valoare exactă și studentul furnizează alegerea corespunzătoare care reflectă corect această valoare în conformitate cu contextul problemei, tratează-o de asemenea ca fiind corectă.\\
4. **Echivalență Numerică**: Tratează răspunsurile numerice ca fiind echivalente dacă sunt corecte cu cel puțin două zecimale sau mai mult, în funcție de precizia furnizată în soluție. De exemplu, atât 0.913, cât și 0.91 ar trebui acceptate dacă soluția este exactă cu două zecimale.\\
5. **Identități Algebrice și Simbolice**: Recunoaște și acceptă forme algebrice echivalente, cum ar fi $\backslash$( $\backslash$sin\string^2(x) + $\backslash$cos\string^2(x) = 1 $\backslash$) sau $\backslash$( e\string^\{i$\backslash$pi\} + 1 = 0 $\backslash$), ca fiind corecte.\\
6. **Forme Trigonometrice și Logaritmice**: Acceptă expresii trigonometrice și logaritmice echivalente, recunoscând identități și transformări care ar putea modifica forma, dar nu și valoarea.\\
7. **Demonstrații Matematice**: Evaluează demonstrațiile matematice pe baza corectitudinii și a logicii, nu a stilului sau a formei. Asigură-te că demonstrațiile sunt complete și corecte, chiar dacă sunt prezentate într-un mod diferit de soluția furnizată.\\

\#\# Formatul Așteptat al Răspunsului: Prezintă răspunsul final cu un scor doar de '0' sau '1', unde '0' semnifică o soluție greșită, iar '1' semnifică o soluție corectă. Nu include nicio altă informație sau explicații suplimentare în răspuns.\\

Problema de matematică este:\\
\{question\}.\\

Soluția corectă din baremul de corectare este:\\
\{true\}.\\

Te rog să evaluezi soluția studentului cu precizie pentru a asigura o evaluare exactă și corectă."""\\
    \},\\
    \{\\
        "role": "user", "content": "Soluția studentului este \{prediction\}. Furnizează un doar scor de '0' sau '1', unde '0' semnifică o soluție greșită, iar '1' semnifică o soluție corectă. Bazează-ți evaluarea pe criteriile de evaluare furnizate si pe soluția corecta din barem.",\\
    \}\\

    \end{tcolorbox}
    \caption{Romanian judge prompt.}
    \label{tab:romanian-judge-prompt}
\end{table*}
\begin{table*}[hbt!]
    \begin{tcolorbox}[
        enhanced,
        attach boxed title to top left={xshift=6mm,yshift=-3mm},
        colback=white,
        colframe=gray!20,
        colbacktitle=white,
        fonttitle=\bfseries\color{black},
        boxed title style={size=small,colframe=gray!20,rounded corners},
        rounded corners,
        ]
        \vspace{2mm}
        \small

    \{\\
        "role": "system",\\
        "content": """Assume the role of a math teacher responsible for evaluating student responses for a math problem against the provided correct solutions. Solutions may include proofs, exact values, multiple-choice answers, or numerical approximations.\\

\#\# Evaluation Criteria:\\
1. **Mathematical Equivalence**: Evaluate answers based on mathematical equivalence, not just numerical accuracy. Check if different algebraic or symbolic expressions are equivalent. Ensure that there are equivalences such as $\backslash$( $\backslash$frac\{$\backslash$sqrt\{6\}-$\backslash$sqrt\{2\}\}\{2\} $\backslash$) being equivalent to $\backslash$( $\backslash$sqrt\{2 - $\backslash$sqrt\{3\}\} $\backslash$).\\
2. **Scoring**: Assign a score of '1' for any answer that matches or is equivalent to the provided solution, whether it is an exact value, a choice label (e.g., A, B, C), or a correctly rounded numerical approximation. Assign a score of '0' for incorrect answers. Do not provide any explanatory feedback.\\
3. **Handling Multiple Choices**: If the solution provided is a choice (e.g., A, B, C, D, E, F) and the student identifies this choice correctly, treat it as correct. If the solution is an exact value and the student provides the corresponding choice that reflects this value correctly according to the problem's context, also treat it as correct.\\
4. **Numerical Equivalence**: Treat numerical answers as equivalent if they are correct to at least two decimal places or more, depending on the precision provided in the solution. For instance, both 0.913 and 0.91 should be accepted if the solution is accurate within two decimal places.\\
5. **Symbolic and Algebraic Identities**: Recognize and accept equivalent algebraic forms, such as $\backslash$( $\backslash$sin\string^2(x) + $\backslash$cos\string^2(x) = 1 $\backslash$) or $\backslash$( e\string^\{i$\backslash$pi\} + 1 = 0 $\backslash$), as correct.\\
6. **Trigonometric and Logarithmic Forms**: Accept equivalent trigonometric and logarithmic expressions, acknowledging identities and transformations that might alter the form but not the value.\\
7. **Mathematical Proofs**: Evaluate mathematical proofs based on correctness and logic, not style or form. Ensure proofs are complete and correct, even if presented differently from the provided solution.
\\
\#\# Expected Output Format: Present your final answer with a score of '0' or '1' only, where '0' signifies an incorrect solution, and '1' signifies a correct solution. Do not include any additional information or feedback in your response.\\

The math problem is:\\
\{question\}.\\

The correct solution from the answer key is:\\
\{true\}.\\

Please evaluate the student's response with precision to ensure accurate and fair grading."""\\
    \},\\
    \{\\
        "role": "user", \\
        "content": "The student answer is {prediction}. Provide a score of '0' or '1' only, where '0' signifies an incorrect solution, and '1' signifies a correct solution. Base your evaluation on the provided evaluation criteria and the correct solution from the answer key.",\\
    \}
    \end{tcolorbox}
    \caption{English judge prompt.}
    \label{tab:english-judge-prompt}
\end{table*}

\begin{table*}[hbt!]
    \begin{tcolorbox}[
        enhanced,
        attach boxed title to top left={xshift=6mm,yshift=-3mm},
        colback=white,
        colframe=gray!20,
        colbacktitle=white,
        fonttitle=\bfseries\color{black},
        boxed title style={size=small,colframe=gray!20,rounded corners},
        rounded corners,
        ]
        \vspace{2mm}
        \tiny
        \textbf{\underline{RoMath-Baccalaureate}} \\
        
        \textbf{\underline{deepseek-ai/deepseek-math-7b-instruct}}\\
        \textbf{\textit{Problem Statement}:} Să se calculeze modulul numărului complex \(\sqrt{3}+i\).\\
        \textbf{\textit{Solution}:} 2 \\
        \textbf{\textit{Generated Solution}:} Pentru a calcula modulul unui număr complex, folosim formula \(|z|=\sqrt{a^2+b^2}\), unde \(z=a+bi\).

        În cazul nostru, \(a=\sqrt{3}\) și \(b=1\).
        
        Aplicând formula, obținem:
        
        \(|z|=\sqrt{(\sqrt{3})^2+1^2}=\sqrt{3+1}=\sqrt{4}=2\).
        
        Deci, modulul numărului complex \(\sqrt{3}+i\) este 2.
        
        Răspuns: \(\boxed{2}\).\\\\

        \textbf{\underline{OpenLLM-Ro/RoLlama3-8b-Instruct}}\\
        \textbf{\textit{Problem Statement}:} Să se demonstreze că vectorii \(\vec{u}=-2 \vec{i}+3 \vec{j}\) și \(\vec{v}=3 \vec{i}+2 \vec{j}\) sunt perpendiculari.\\
        \textbf{\textit{Solution}:} \(\operatorname{Din} \vec{a} \cdot \vec{b}=-2 \cdot 3+3 \cdot 2=0 \Rightarrow \vec{a} \perp \vec{b}\) \\
        \textbf{\textit{Generated Solution}:} Pentru a demonstra că vectorii \(\vec{u}=-2 \vec{i}+3 \vec{j}\) și \(\vec{v}=3 \vec{i}+2 \vec{j}\) sunt perpendiculari, trebuie să arătăm că produsul lor punct este egal cu 0.

Produsul punct al celor doi vectori \(\vec{u}\) și \(\vec{v}\) este dat de:

\(\vec{u} \cdot \vec{v} = (-2 \vec{i} + 3 \vec{j}) \cdot (3 \vec{i} + 2 \vec{j})\)

Acum, putem calcula produsul punct:

\(\vec{u} \cdot \vec{v} = (-2)(3) + (3)(2) = -6 + 6 = 0\)

Deoarece produsul punct al celor doi vectori este egal cu 0, vectorii \(\vec{u}\) și \(\vec{v}\) sunt perpendiculari.\\\\

        \textbf{\underline{RoMath-Competitions}} \\
        
        \textbf{\underline{deepseek-ai/deepseek-math-7b-instruct}}\\
        \textbf{\textit{Problem Statement}:} Într-o clasă sunt 22 de elevi, dintre care 12 sunt fete. Să se determine în câte moduri se poate alege un comitet al clasei format din 3 fete și 2 băieți.\\
        \textbf{\textit{Solution}:} Numărul băieților este \(22-10=12\). Fetele pot fi alese în \(\mathrm{C}_{12}^{3}=220\) moduri. Băieții pot fi aleși în \(C_{10}^{2}=45\) moduri. Comitetul format din 3 fete și 2 băieți dintre cei 22 de elevi ai clasei poate fi format în \(\mathrm{C}_{12}^{3} \cdot \mathrm{C}_{10}^{2}=9900\) moduri. \\
        \textbf{\textit{Generated Solution}:} Pentru a rezolva această problemă, trebuie să utilizăm formula combinării. Combinarea este o modalitate de a alege un set de elemente dintr-o mulțime mai mare, fără a uita la ordinea elementelor.

În acest caz, trebuie să alegem 3 fete din 12 și 2 băieți din 10 (deoarece 22 - 12 = 10).

Formula combinării este:

$$C(n, k) = \frac{n!}{k!(n-k)!}$$

unde n este numărul total de elemente, k este numărul de elemente ce trebuie alese, iar ""!"" semnifică factorialul unui număr.

Pentru a calcula numărul de moduri în care putem alege 3 fete din 12, aplicăm formula:

$$C(12, 3) = \frac{12!}{3!(12-3)!} = \frac{12!}{3!9!} = \frac{12 \times 11 \times 10}{3 \times 2 \times 1} = 220$$

Pentru a calcula numărul de moduri în care putem alege 2 băieți din 10, aplicăm formula:

$$C(10, 2) = \frac{10!}{2!(10-2)!} = \frac{10!}{2!8!} = \frac{10 \times 9}{2 \times 1} = 45$$

Pentru a calcula numărul total de moduri în care putem forma comitetul, înmulțim cele două rezultate:

$$220 \times 45 = 9900$$

Deci, există 9900 de moduri în care se poate forma comitetul.

Răspunsul final este $\boxed{9900}$.\\\\

        \textbf{\underline{OpenLLM-Ro/RoLlama3-8b-Instruct}}\\
        \textbf{\textit{Problem Statement}:} Un grădinar a plantat într-una dintre grădinile pe care le îngrijește parcele cu tufe de trandafir din soiuri distincte, astfel încât fiecare parcelă conține trandafiri dintr-un alt soi. Într-o zi are de realizat un aranjament floral din acești trandafiri. Analizează tufele și procedează în felul următor: din prima parcelă taie trei trandafiri, din cea de a doua taie de două ori mai mulți decât din prima ș.a.m.d., tăind dintr-o parcelă de două ori mai mulți trandafiri decât din parcela precedentă. Care este numărul minim de trandafiri pe care trebuie să-1 aibă parcela a IX-a pentru a putea fi folosită la realizarea aranjamentului floral?\\
        \textbf{\textit{Solution}:} Parcela a IX a trebuie să aibă cel puțin: \(3 \cdot 2^{8}=3 \cdot 256=768\) \\
        \textbf{\textit{Generated Solution}:} Să notăm numărul de trandafiri din fiecare parcelă ca P1, P2, P3,..., P9. Conform problemei, numărul de trandafiri din fiecare parcelă este dat de:

P1 = 3\\
P2 = 2 * P1 = 2 * 3 = 6\\
P3 = 2 * P2 = 2 * 6 = 12\\
P4 = 2 * P3 = 2 * 12 = 24\\
P5 = 2 * P4 = 2 * 24 = 48\\
P6 = 2 * P5 = 2 * 48 = 96\\
P7 = 2 * P6 = 2 * 96 = 192\\
P8 = 2 * P7 = 2 * 192 = 384\\
P9 = 2 * P8 = 2 * 384 = 768

Deci, numărul minim de trandafiri din parcela a IX-a este 768.\\\\
    
    \end{tcolorbox}
    \caption{Qualitative examples of correct zero-shot predictions for RoMath-Baccalaureate and RoMath-Competitions.}
    \label{tab:correct-predictions}
\end{table*}
\begin{table*}[hbt!]
    \begin{tcolorbox}[
        enhanced,
        attach boxed title to top left={xshift=6mm,yshift=-3mm},
        colback=white,
        colframe=gray!20,
        colbacktitle=white,
        fonttitle=\bfseries\color{black},
        boxed title style={size=small,colframe=gray!20,rounded corners},
        rounded corners,
        ]
        \vspace{2mm}
        \tiny
        \textbf{\underline{RoMath-Baccalaureate}} \\

        \textbf{\underline{deepseek-ai/deepseek-math-7b-instruct}}\\
        \textbf{\textit{Problem Statement}:} Să se rezolve în mulțimea numerelor reale ecuația \(2^{x}=4^{x}\).\\
        \textbf{\textit{Solution}:} x=0 \\
        \textbf{\textit{Generated Solution}:} Pentru a rezolva ecuația \(2^{x}=4^{x}\), putem scrie \(4^{x}\) ca \((2^{2})^{x}\), adică \(2^{2x}\).

Astfel, ecuația devine \(2^{x}=2^{2x}\).

Prin urmare, putem scrie \(x=2x\).

Împărțind ambii membri ai ecuației cu \(x\), obținem \(1=2\).

Această ecuație nu are soluții în mulțimea numerelor reale, deoarece 1 nu este egal cu 2.

Deci, ecuația \(2^{x}=4^{x}\) nu are soluții în mulțimea numerelor reale.

Răspuns: \(\boxed{$Nu are soluții$}\) \\\\

        \textbf{\underline{OpenLLM-Ro/RoLlama3-8b-Instruct}}\\
        \textbf{\textit{Problem Statement}:} Să se calculeze determinantul \(\left|\begin{array}{ccc}1 & 3 & 5 \\ 7 & 9 & 11 \\ 2 & 4 & 6\end{array}\right|\). \\
        \textbf{\textit{Solution}:} 0 \\
        \textbf{\textit{Generated Solution}:} Pentru a calcula determinantul matricei date, putem folosi metoda de expansiune prin rânduri. Să calculăm determinantul:

        \(\left|\begin{array}{ccc}1 & 3 & 5 \\ 7 & 9 & 11 \\ 2 & 4 & 6\end{array}\right|\)
        
        În primul rând, putem elimina primul rând și primul element din al doilea rând:
        
        \(\left|\begin{array}{cc}9 & 11 \\ 2 & 4\end{array}\right|\)
        
        Acum, putem calcula determinantul matricei 2x2 rămase:
        
        \(\left|\begin{array}{cc}9 & 11 \\ 2 & 4\end{array}\right| = 9 \cdot 4 - 11 \cdot 2 = 36 - 22 = 14\)
        
        Deci, determinantul matricei date este 14.\\\\

        \textbf{\underline{RoMath-Competitions}} \\
        
        \textbf{\underline{deepseek-ai/deepseek-math-7b-instruct}}\\
    
        \textbf{\textit{Problem Statement}:} Într-un plan, raportat la reperul ortogonal de axe de coordonate \((x \mathrm{xOy})\) se dau punctele: \(\mathrm{A}(0 ; 6)\); \(\mathrm{B}(5 ; 4) ; \mathrm{C}(-1 ; 4)\). Pentru \(\mathrm{a}=5\) să se scrie ecuația medianei corespunzătoare laturii \(\mathrm{BC}\). \\
        \textbf{\textit{Solution}:} Determină coordonatele mijlocului lui \([\mathrm{BC}]\) notat cu \(\mathrm{M}\) :
        \(\mathrm{x}_{\mathrm{M}}=\frac{5-1}{2}=2, \mathrm{y}_{\mathrm{M}}=\frac{4+4}{2}=4 \Leftrightarrow \mathrm{M}(2,4)\)
        Ecuația dreptei (AM): \(\left|\begin{array}{lll}x & y & 1 \\ 0 & 6 & 1 \\ 2 & 4 & 1\end{array}\right|=0 \Rightarrow A M: x+y-6=0\). \\
        \textbf{\textit{Generated Solution}:} Pentru a scrie ecuația medianei corespunzătoare laturii BC, trebuie să găsim mijlocul segmentului BC, notat M.

Mijlocul unui segment este dat de media aritmetică a coordonatelor punctelor de capăt ale segmentului. Astfel, coordonatele punctului M sunt:

M(xM; yM) = ((xB + xC) / 2; (yB + yC) / 2)

M(-1 + 5 / 2; 4 + 4 / 2) = M(2; 4)

Mediana AM este dreapta care unește punctul A cu mijlocul M al segmentului BC.

Ecuația unei drepte în plan, cunoscând coordonatele a două puncte A(x1; y1) și B(x2; y2), este: (y - y1) / (y2 - y1) = (x - x1) / (x2 - x1)

Aplicând formula pentru mediana AM, avem: (y - 6) / (4 - 6) = (x - 0) / (2 - 0)

Simplificând, obținem: (y - 6) / (-2) = x / 2

Ecuația medianei AM este: y - 6 = -2x sau 2x + y - 6 = 0

Pentru a = 5, ecuația medianei corespunzătoare laturii BC este: 2x + y - 6 = 0\\

        \textbf{\underline{OpenLLM-Ro/RoLlama3-8b-Instruct}}\\
        \textbf{\textit{Problem Statement}:} În raport cu un reper cartezian \(x O y\), considerăm un purice \(P\) care sare doar în puncte având ambele coordonate întregi. La o săritură, puricele se deplasează doar pe verticală sau pe orizontală. După o săritură nu este obligatoriu să schimbe direcția de deplasare, dar respectă următoarea regulă: sare 3 unități, apoi 2 unități, 3 unități, 2 unităti etc. De exemplu, dacă M se află inițial în punctul \((1,2)\), la prima mutare ar putea fi în punctul \((1,5)\), iar apoi în \((1,7)\) dacă păstrează direcția, sau în ( 3,5 ), dacă își schimbă direcția. Dacă \(\mathrm{P}\) se află inițial în origine, demonstrați că poate ajunge în punctul \((2013,0)\).\\
        \textbf{\textit{Solution}:} Puricele se poate deplasa doar orizontal, astfel: după două sărituri ajunge în punctul \((5,0)\), după patru ajunge în \((10,0)\) și, tot așa, după 804 sărituri ajunge în \((2010,0)\). La următoarea săritură va ajunge în \((2013,0)\). \\
        \textbf{\textit{Generated Solution}:} Pentru a demonstra că puricele poate ajunge în punctul \((2013,0)\), vom arăta că poate ajunge în orice punct cu coordonate întregi prin repetarea ciclului de deplasare.

Să notăm deplasarea verticală ca \(v\) și deplasarea orizontală ca \(h\). Ciclul de deplasare este dat de secvența \((v, h, v, h, v, h, \dots)\), unde \(v = 3\) și \(h = 2\). Aceasta înseamnă că puricele se deplasează în sus 3 unități, apoi în dreapta 2 unități, apoi în sus 3 unități, apoi în dreapta 2 unități, și așa mai departe.

Acum, să luăm în considerare punctul \((2013,0)\). Deoarece puricele poate sări doar în puncte cu coordonate întregi, trebuie să găsim o secvență de deplasări care să conducă la acest punct. Putem observa că ciclul de deplasare dat poate fi repetat de 2013 ori pentru a ajunge la punctul \((2013,0)\).

Pentru a vedea acest lucru, să luăm în considerare deplasările necesare pentru a ajunge la punctul \((2013,0)\) din origine:

1. Deplasare verticală de 3 unități (de la \((0,0)\) la \((0,3)\)).

2. Deplasare orizontală de 2 unități (de la \((0,3)\) la \((2,3)\)).

3. Deplasare verticală de 3 unități (de la \((2,3)\) la \((2,6)\)).

4. Deplasare orizontală de 2 unități (de la \((2,6)\) la \((4,6)\)).

5. Deplasare verticală de 3 unități (de la \((4,6)\) la \((4,9)\)).

6. Deplasare orizontală de 2 unități (de la \((4,9)\) la \((6,9)\)).

7. Deplasare verticală de 3 unități (de la \((6,9)\) la \((6,12)\)).

8. Deplasare orizontală de 2 unități (de la \((6,12)\) la \((8,12)\)).

9. Deplasare verticală de 3 unități (de la \((8,12)\) la \((8,15)\)).

10. Deplasare orizontală de 2 unități (de la \((8,15)\) la \((10,15)\)).

Putem continua acest proces până când ajungem la punctul \((2013,0)\). Deoarece ciclul de deplasare poate fi repetat de 2013 ori, puricele poate ajunge în punctul \((2013,0)\) după efectuarea ciclului de deplasare de 2013 ori.\\\\
    
    \end{tcolorbox}
    \caption{Qualitative examples of incorrect zero-shot predictions for RoMath-Baccalaureate and RoMath-Competitions.}
    \label{tab:incorrect-predictions}
\end{table*}

\end{document}